\documentclass[1p,times]{elsarticle}
\usepackage{graphicx}
\usepackage{color}
\usepackage{amssymb}
\usepackage{amsmath}
\usepackage{xspace}
\usepackage{multirow}   
\usepackage{booktabs}   
\usepackage{longtable}
\usepackage{xcolor}

\usepackage{hyperref}
\hypersetup{colorlinks=true, linkcolor=black, citecolor=black, urlcolor=black}
\usepackage{lineno}

\biboptions{comma, square, sort&compress}
\journal{Communications in Nonlinear Science and Numerical Simulation} 
\begin{document}
\begin{frontmatter}



\ead{thomas.schrefl@donau-uni.ac.at}
\cortext[cor1]{Corresponding author} 

\title{Conditional physics informed neural networks}
\author[a,b]{Alexander Kovacs}
\author[c,d,e]{Lukas Exl}
\author[a,b]{Alexander Kornell}
\author[a,b]{Johann Fischbacher}
\author[a,b]{Markus Hovorka}
\author[a,b]{Markus Gusenbauer}
\author[b]{Leoni Breth}
\author[b]{Harald Oezelt}
\author[f]{Masao Yano}
\author[f]{Noritsugu Sakuma}
\author[f]{Akihito Kinoshita}
\author[f]{Tetsuya Shoji}
\author[f]{Akira Kato}
\author[a,b]{Thomas Schrefl\corref{cor1}}

\address[a]{Christian Doppler Laboratory for Magnet design through physics informed machine learning, Viktor Kaplan-Stra{\ss}e 2E, 2700 Wiener Neustadt, Austria}
\address[b]{Department for Integrated Sensor Systems, Danube University Krems, Viktor Kaplan-Stra{\ss}e 2E, 2700 Wiener Neustadt, Austria}
\address[c]{Department of Mathematics, University of Vienna, Oskar-Morgenstern-Platz 1, 1090 Vienna, Austria}
\address[d]{Wolfgang Pauli Institute, Oskar-Morgenstern-Platz 1, 1090 Vienna, Austria}
\address[e]{Research Platform MMM Mathematics-Magnetism-Materials, Oskar-Morgenstern-Platz 1, 1090 Vienna, Austria}
\address[f]{Advanced Materials Engineering Div., Toyota Motor Corporation, 1200, Mishuku Susono, Shizuoka 410-1193 Japan}


\begin{abstract}
We introduce conditional PINNs (physics informed neural networks) for estimating the solution of classes of eigenvalue problems. The concept of PINNs is expanded to learn not only the solution of one particular differential equation but the solutions to a class of problems. We demonstrate this idea by estimating the coercive field of permanent magnets which depends on the width and strength of local defects. When the neural network incorporates the physics of magnetization reversal, training can be achieved in an unsupervised way. There is no need to generate labeled training data. The presented test cases have been rigorously studied in the past. Thus, a detailed and easy comparison with analytical solutions is made. We show that a single deep neural network can learn the solution of partial differential equations for an entire class of problems.   
\end{abstract}

\end{frontmatter}


\section{\label{sec:introduction}Introduction}

Neural networks have been widely used to estimate the solution of partial differential equations \cite{khan2019deep,kim2019deep,kovacs2019learning}. A common approach has two steps. Firstly, training data is generated by solving the partial differential equation with finite element, finite volume, or finite difference solvers. Data may be generated for different scenarios like different source terms or boundary conditions, 
for progressing the solution in time, or a combination of both. Secondly, the generated data is used to train a neural network, which in turn is used to quickly estimate physical fields. As the network has been trained for different scenarios, approximate solutions of the partial differential equation can be obtained without the need for meshing and numerical solution. Khan and co-workers \cite{khan2019deep} applied this approach for Maxwell's equation to estimate the magnetic field in magnetic machines. Kim and co-workers \cite{kim2019deep}  trained a neural network for solutions of the in-compressible Navier-Stokes equation in order to estimate the time evolution of smoke clouds for different source terms. Kovacs and co-workers \cite{kovacs2019learning} used neural networks for the approximation of the time evolution of the magnetization according to the Landau-Lifshitz-Gilbert equation. For these examples, the weights of the neural network are found by minimizing a loss function, which is related to the difference between the numerical solution of the partial differential equation and the neural network approximation of the solution. For approximating valid solutions, parts of the physics were included in the loss function. Kim and co-workers \cite{kim2019deep} used a special loss function that guarantees a divergence-free velocity field. Kovacs and co-workers \cite{kovacs2019learning} included the constraint that the norm of magnetization remains constant as penalty term in the loss function.    

Physics informed neural networks \cite{raissi2019physics} include the physics of the underlying problem in the loss function. The loss function of physics informed neural networks, which is minimized during training, is directly computed from the governing partial differential equation. The loss function is either formed by the residuals at collocation points \cite{koryagin2019pydens}, the weighted residuals obtained by the Galerkin-Method \cite{kharazmi2019variational}, or the energy functional of an Euler-Lagrange differential equation \cite{e2018deep}. Thus, for training a physics informed neural network there is no need to generate labeled training data in advance. The input data for physics informed neural networks are points sampled in the problem domain. The loss function can be augmented with the distance between the approximated solution of the partial differential equation and measured values of the solution. Then one or more coefficients of the partial differential equation can be included as unknowns during training. In this way an inverse problem is solved. Solving inverse problems with physics informed neural networks may lead to a significant speed up as compared to conventional methods \cite{hennigh2020nvidia}. Physics informed neural networks can be used to solve eigenvalue problems, when the loss function of a neural network is related to the Rayleigh-Ritz coefficient \cite{e2018deep}.      

Micromagnetism \cite{brown1963micromagnetics} is a continuum theory that describes magnetization processes at a length scale that is large enough to replace discrete atomic moments with a continuous function of space and small enough to resolve magnetic domains. Traditionally, the Ritz method has been applied to solve micromagnetic problems numerically. An ansatz for the magnetization was made in terms of one or more free variables. The unknown coefficients for the magnetization field were found by minimization of the Gibbs free energy of the systems. Brown \cite{brown1957criterion} used the Ritz method to derive the switching field of a ferromagnetic cylinder. Kondorksy \cite{Kondorsky1979stability} estimated the hysteresis loops of fine ferromagnetic particles, using the Ritz method.  Finite element micromagnetic solvers \cite{fredkin1987numerical,schrefl1994two} use the same approach. The magnetization is expanded in terms of the basis functions on a tetrahedral finite element mesh.   

In this work we show that a physics informed neural network can be used to solve not only a single unique problem but also a class of eigenvalue problems. Using the parameters that determine the coefficients of an eigenvalue equation as conditional input to the neural network, the network can be trained to approximate the solutions of a class of partial differential equations spanned by the set of parameters. We apply conditional physics informed neural networks to solve eigenvalue problems in micromagnetics. The eigenvalue is the critical field when a magnet starts to reverse under the influence of an external field. In particular, we focus on classical problems in micromagnetics for which the solutions for the nucleation field are well known. Thus, we can directly compare estimates for nucleation fields obtained by physics informed neural networks with the analytic solutions.   

\section{Neural networks for solving eigenvalue problems}
\subsection{Variational form of eigenvalues problems}\label{sec:varform}
Following Komzsik \cite{komzsik2019applied}, we show how the smallest eigenvalue and the corresponding eigensolution of  the Sturm-Liouville eigenproblem can be found using variations. For readability, we restrict ourselves to the one-dimensional case. We start with differential equation  
\begin{align}
	\label{eq:sturmliouville}
	\frac{\mathrm{d}}{\mathrm{d}x} \left( p(x) \frac{\mathrm{d}y}{\mathrm{d}x}\right) + q(x)y(x) = -\lambda r(x)y(x).
\end{align}
The unknown solution function $y(x)$ is the eigensolution and $\lambda$ is the eigenvalue. The left-hand side of (\ref{eq:sturmliouville}) is the Sturm-Liouville operator. 
In classical theory the known functions $p(x) > 0$, $q(x)$, and $r(x) > 0$ are continuous and continuously differentiable, however, in the setting of variational treatment of elliptic 
eigenvalue problems  
it suffices that the coefficients are bounded together with $r(x),p(x)$ being positive (which generalizes to an ellipticity assumption in higher dimensions) \cite{henrot2006extremum}. 
For now, we assume homogeneous Dirichlet boundary conditions
\begin{align}
	\label{eq:boundary}
	y(x_1) = 0,\; y(x_2) = 0.
\end{align}
Inhomogeneous boundary conditions can be turned into this case by a simple redefinition of $y$.
The eigensolution for the smallest eigenvalue can be found by minimizing the functional
\begin{align}
	\label{eq:functional}
	I(y) = \int_{x_1}^{x_2} \left\{ p(x) \left(\frac{\mathrm{d}y(x)}{\mathrm{d}x}\right)^2 - q(x) y^2(x)\right\}\mathrm{d}x \rightarrow \mathrm{min}
\end{align} 
subject to the constraint
\begin{align}
	\label{eq:constraint}
	\int_{x_1}^{x_2} r(x)y^2(x) \mathrm{d}x = 1
\end{align} 
and the conditions $y(x_1) = 0$ and $y(x_2) = 0$.
The smallest eigenvalue is obtained by minimizing the Rayleigh quotient  
\begin{align}
	\lambda = {\min_{y}} \frac{\int_{x_1}^{x_2} \left\{ p(x) \left(\frac{\mathrm{d}y(x)}{\mathrm{d}x}\right)^2 - q(x) y^2(x)\right\}\mathrm{d}x}{\int_{x_1}^{x_2} r(x)y^2(x) \mathrm{d}x}
\end{align} 

\subsection{Ritz and Kantorovich methods}
\label{sec:ritztest}
Within the framework of the Ritz method, the unknown solution $y(x)$ is expressed as a linear combination of basis functions $f_i$. The basis functions are chosen in such a way that they fulfill the boundary conditions (\ref{eq:boundary}). Plugging the approximate solution
\begin{align}
   y_\mathrm{approx}(x) = \sum_i c_i f_i(x)
\end{align}	
into (\ref{eq:functional}) and (\ref{eq:constraint}) leads to a constraint algebraic minimization problem for the coefficients $c_i$. For homogeneous boundary conditions $y(x_1) = 0$ and $y(x_2)=0$ the basis functions need to vanish at the boundary.  Kantorovich \cite{kantorovich1958approximate} suggested constructing such basis functions as 
\begin{align}
\label{eq:trick}
f_i(x) = w(x) g_i(x)
\end{align}
whereby the function $w(x)$ fulfills $w(x) \ge 0$ for $x_1 < x < x_2$ and $w(x) = 0$ for $x = x_1$ or $x = x_2$. This ansatz relaxes the restrictions on the basis functions allowing great flexibility for the functions $g_i(x)$.  
Another choice is to apply natural boundary conditions. As shown by Gould \cite{gould2012variational},  minimizing the functional (\ref{eq:functional}) without imposing a prescribed boundary condition gives a minimizer $y(x)$ that solves the eigenvalue problem (\ref{eq:sturmliouville}) 
and fulfills the natural boundary conditions $\mathrm{d}y(x_1)/\mathrm{d}x = 0$ and $\mathrm{d}y(x_2)/\mathrm{d}x = 0$. 

One particular choice is the approximation of the unknown solution with a dense neural network.  We either apply Kantorvich's trick (\ref{eq:trick}) and approximate the unknown solution by 
\begin{align}
	\label{eq:nn_dirichlet}
	y_\mathrm{approx}(x) = w(x) \mathcal{N}(x,\mathbf{w}),
\end{align}
for problems for which the solution is zero at the boundary or simply use
\begin{align}
	\label{eq:nn_neumann}
	y_\mathrm{approx}(x) = \mathcal{N}(x,\mathbf{w}),
\end{align}
for problems with vanishing normal derivative of the solution at the boundary. Here
$\mathcal{N}(x,\mathbf{w})$ is the output of a dense multi-layer neural network with input $x$. The vector $\mathbf{w}$ represents the weights and biases of the network. The weights and biases are the learnable parameters of the network which are determined during training of the network by minimizing the functional \begin{align}
	\label{eq:loss}
	L(y_\mathrm{approx}(x)) = \frac{\int_{x_1}^{x_2} \left\{ p(x) \left(\frac{\mathrm{d}y_\mathrm{approx}(x)}{\mathrm{d}x}\right)^2 - q(x) y_\mathrm{approx}^2(x)\right\}\mathrm{d}x}{\int_{x_1}^{x_2} r(x)y_\mathrm{approx}^2(x) \mathrm{d}x} + \gamma \left(\int_{x_1}^{x_2} r(x) y_\mathrm{approx}^2(x)\mathrm{d}x - 1 \right)^2
\end{align}  
with a variant of the stochastic gradient descent algorithm. The second term on the right-hand side is a penalty term that accounts for the constraint (\ref{eq:constraint}).
As discussed by E and Yu \cite{e2018deep} using both the denominator in the first term and the penalty term relaxes the choice of the parameter $\gamma$ and improves the convergence. Please note that within the framework of the conventional Ritz method two alternatives are well-known. (1) The first term of (\ref{eq:loss}) corresponds to the Rayleigh quotient and was minimized by Skomski \cite{skomski2002exact} for computing nucleation fields in ferromagnetic wires. (2) Using (\ref{eq:loss}) without the denominator in the first term was applied by Komzsik \cite{komzsik2019applied}.
During training neural network algorithms we evaluate the integrals in (\ref{eq:loss}) by a quasi-Monte Carlo method. For example, we approximate the integral
\begin{align}
	\int_{x_1}^{x_2} F(x) dx \approx \frac{x_2-x_1}{N} \sum_i F(x_i),
\end{align}
where $N$ is the batch-size of the gradient descent algorithm and the points $x_i$ are sampled from the interval $[x_1,x_2]$ using a quasi-random sequence. The use of quasi-random points for Monte-Carlo integration improves convergence \cite{caflisch1998monte} since clumps of points that occur for random sampling can be avoided. In particular, we apply the Sobol sequence as implemented in the Python library scikit-optimize \cite{head10scikit}. Similarly, Hennig and co-workers \cite{hennigh2020nvidia} apply quasi-Monte Carlo integration to evaluate the integrals occurring during the solution of partial differential equations with physics informed neural networks.

\subsection{Learning solutions for classes of eigenvalue problems}

The input for the neural network (\ref{eq:nn_dirichlet}) or (\ref{eq:nn_neumann}) is a point in space. For a batch of points the function (\ref{eq:loss}) is evaluated. During training the neural network receives several batches and the weights are adjusted by minimization of (\ref{eq:loss}). 
Typically, a neural network is trained for a particular choice of $p(x)$, $q(x)$, and  $r(x)$ to solve a forward problem for a given specific differential equation. 
Also, physics informed neural networks have been successfully applied to inverse problems. 
Given additional constraints, unknown coefficients of the differential equations are used as free parameters during minimization of the loss function \cite{koryagin2019pydens,lu2021deepxde,haghighat2021sciann}.     

\begin{figure}[!htb]
	\centering
	\includegraphics[scale=0.5]{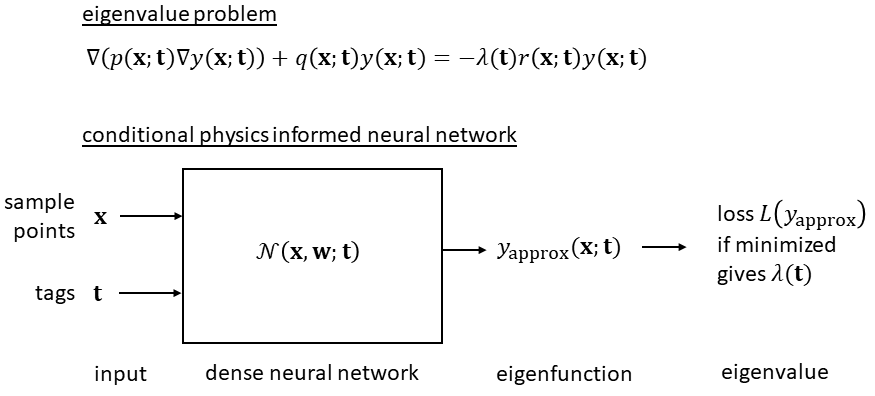}
	\caption{\label{fig:cpinn}Schematics of a conditional physics informed neural network for the solution of classes of three-dimensional eigenvalue problems. The coefficients of the eigenvalue equation may depend on a vector of parameters $\mathbf{t}$. The inputs of the network are the sample points and the tag vectors. The network approximates the eigenfunction for whole space parameters which is covered by the training data. The tag vector serves as conditional input that selects a specific eigenvalue problem.} 
\end{figure}

\begin{figure}[!htb]
	\centering
	\includegraphics[scale=0.5]{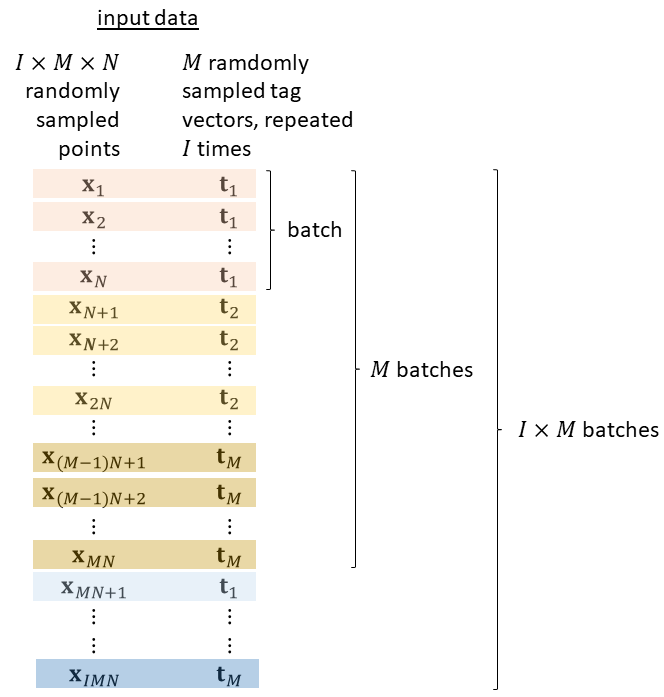}
	\caption{\label{fig:data}Input data for a conditional physics informed neural network. Each input record consists of a point sampled from the problem domain $\mathbf{x}_i$ and a vector of tags $\mathbf{t}_j$. $N$ input records form a batch. All records within a batch have the same tag vector. The parameter space is sampled by $M$ tag vectors. The same set of tags is repeatedly used $I$ times. The total number of input records used for training is $I\times M\times N$. In total, $I\times M\times N$ points are randomly sampled from the problem domain. The number of batches is $I\times M$.} 
\end{figure}

Here we introduce another area where physics informed neural networks might be useful. When solving forward problems, we have the freedom to provide additional input to the neural network and thus expand the scope of the learning. Suppose one or more of the functions $p(x)$, $q(x)$, or $r(x)$ depend on a vector of parameters $\mathbf{t} = (t_1,t_2,,..)^\mathrm{T}$. We call the components of $\mathbf{t}$ tags. Setting the tags  in $\mathcal{N}(x,\mathbf{w};\mathbf{t})$ selects a specific equation out of a class of problems during training. During prediction, the tags are a conditional input that selects a specific solution. Figure~\ref{fig:cpinn} schematically shows the concept of conditional physics informed neural networks. A similar approach was used by He and Pathak \cite{he2020unsupervised} who trained a neural network for the solution of a class of heat equations. They applied an image gradient based network to minimize the residual of the heat equation while providing spatial distribution of the heat source as conditional input.

Input data for training are randomly sampled points from the problem domain and tag vectors sampled from the parameters space. The input records are combined into batches. In each iteration step, the stochastic gradient descent method adjusts the weights of the neural network according to the samples of one batch. The batch size $N$ is the number of points which are used to evaluate the integrals in (\ref{eq:loss}) by quasi-Monte Carlo integration. Figure \ref{fig:data} shows how the input data is composed of points and tags. All input records within one batch have $N$ different points but the same tag vector. The number of different tag vectors sampled from the parameter space is $M$. Across all input records the set of tag vectors is repeated $I$ times. The total number of different points sampled from the problem domain is $I\times M\times N$. Using the same tag vector in multiple batches ($I > 1$) was found to improve the convergence during training.

We apply conditional physics informed neural networks for solving eigenvalue problems in magnetics. For example, we aim for a single neural network predictor that estimates the nucleation field for different defects in permanent magnets. In the most simple case, the defect is defined by two parameters: The width of the defect and its strength \cite{kronmuller1987theory}. Defect width and defect strength are the tags that change the function $p(x)$ and $q(x)$.  

\subsection{Methodology}

We apply the Keras/Tensorflow wrapper SciAnn \cite{haghighat2021sciann} for implementing conditional physics informed neural networks. 

Neural networks can overfit the training data. In this work, eigenvalue problems are selected by the tags which are inputs to the neural network. Thus, the weights are fitted for the eigenvalue problems that are included in the training data.  Overfitting occurs, when the network predicts the solution of these eigenvalue problems too well and cannot generalize to new problems of the same class. We employ the following two strategies to avoid overfitting. First, we try to keep the learnable parameters in the model low. Reducing the capacity of the network is a simple way to limit overfitting \cite{chollet2018deep}. Second, we apply early stopping. We use 80 percent of the input data for training (training set) and 20 percent for validation (validation set). We stop training when the validation loss is no longer improving. The validation loss is the average of (\ref{eq:loss}) over all batches of the validation set. In particular, we exit training, when the validation loss has stopped decreasing for more than twenty epochs. One epoch is a complete run through the training set. The minimum observed validation loss was used for manual optimization of the network layout. The hyperparameters were selected based on the validation loss only. The analytical solution, which  is available for all problems discussed in this work, was used only for testing.

We monitor the validation loss for tuning the step length of the stochastic gradient descent method during minimization the functional (\ref{eq:loss}). Decreasing the learning rate when a plateau of the validation loss occurs avoids getting stuck in a local minimum \cite{chollet2018deep}.  If we observe no reduction of the validation loss function for ten epochs, we reduce the learning rate by a factor of 0.1. 

\section{Micromagnetic background}

\subsection{Micromagnetic energy}
The state of a magnet is given by the unit vector $\mathbf{m}(x)$ which gives the direction of the magnetization vector $\mathbf{M}(x)$. Within the framework of micromagnetism \cite{brown1963micromagnetics}, $\mathbf{m}$ is treated as a continuous vector field.  The norm of $\mathbf{M}$ is the spontaneous magnetization $M_\mathrm{s} = |\mathbf{M}|$ of the material. We can write $\mathbf{m}(x) = \mathbf{M}/M_\mathrm{s}$. Under the action of an applied field $\mathbf{H}_\mathrm{ext}$ the magnetization configuration changes until a new metastable or stable state is reached. In the grains of a permanent magnet the magnetization is oriented parallel to a preferred crystallographic direction. Because of this strong magneto-crystalline anisotropy, the magnet does not demagnetize even if it is exposed to strong magnetic fields. For practical applications the critial field at which the magnetization starts to deviate from its preferred crystallographic direction is of utmost importance. In classical micromagnetic theory, this field is called nucleation field and follows from the solution of an eigenvalue problem \cite{brown1957criterion,aharoni1958magnetization,kronmuller1987theory,skomski1992nucleation}. The nucleation field can give estimates for the coercive field of the magnet for local defects \cite{kronmuller1987theory} or multi-phase magnets \cite{nieber1989nucleation,skomski1993giant}.

\begin{figure}[!htb]
	\centering
	\includegraphics[scale=0.5]{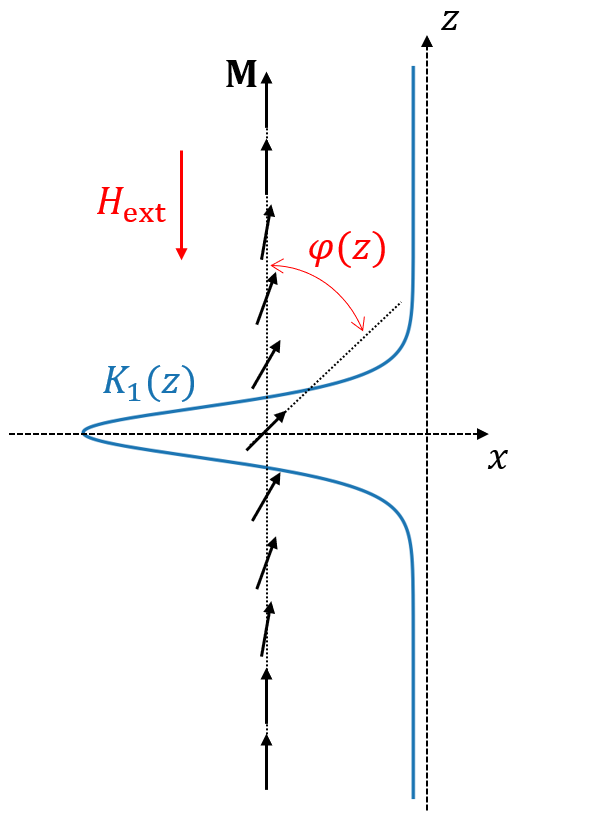}
	\caption{\label{fig:nuc1}Planar defect that infinitely extends in the $x$ and $y$  direction. In the defect, the magnetocrystalline anisotropy $K_1(z)$ is reduced. In this one-dimensional problem, the magnetization angle $\varphi (z)$  depends only on the $z$ direction.} 
\end{figure}

For the purpose of this paper we consider a simple one-dimensional problem which indefinitely extends in the $x$ and $y$ directions. For the generalization please see the classical text books on micromagnetics \cite{brown1963micromagnetics,kronmuller2003micromagnetism,aharoni2000introduction}. A sketch of the one-dimensional micromagnetic problem is given in Figure \ref{fig:nuc1}. With the angle $\varphi(z)$ between $\mathbf{m}$ and the $z$ direction the system can be fully described.  The total Gibbs free energy $E$ is the sum of the exchange energy, the magneto-crystalline anisotropy energy, the Zeeman energy, and the magnetostatic energy. Assuming that the uniaxial magneto-crystalline anisotropy axis coincides with $z$ and the external field $\mathbf{H}_\mathrm{ext}$ is applied in $-z$ direction ($\mathbf{H}_\mathrm{ext} = - H_\mathrm{ext} \hat{\mathbf z}$), we can write the energy
\begin{align}
	\label{eq:energy}
	E = \int \left\{ A \left( \frac{\mathrm{d} \varphi(z) }{\mathrm{d}z} \right)^2 + K_1(z) \sin^2 \varphi(z) + \mu_0 H_\mathrm{ext} M_\mathrm{s} \cos \varphi - \frac{\mu_0}{2} \mathbf{H}_\mathrm{d}(z) \cdot \mathbf{M}(z)\right\} \mathrm{d}z.
\end{align}
Here $A$ is the exchange constant, $K_1$ is the anisotropy constant, and $\mathbf{H}_\mathrm{d}$ is the demagnetizing field, and $\mu_0$ is the permeability of vacuum. The first variation of the energy (\ref{eq:energy}) gives the equilibrium condition for $\varphi(z)$, which will we discuss in the following.

\subsection{Inhomogeneous nucleation in planar defects}
We now assume a planar defect which extends infinitely in the $x$ and $y$ direction. Across the defect region with an extension of $2 r_0$ the anisotropy constant $K_1(z)$ changes from its bulk value. Examples of defects in permanent magnets are the ferromagnetic grain boundary phases \cite{bance2014influence} or soft magnetic inclusions \cite{hirosawa2004development}. 

\begin{figure}[!htb]
	\centering
	\includegraphics[scale=0.5]{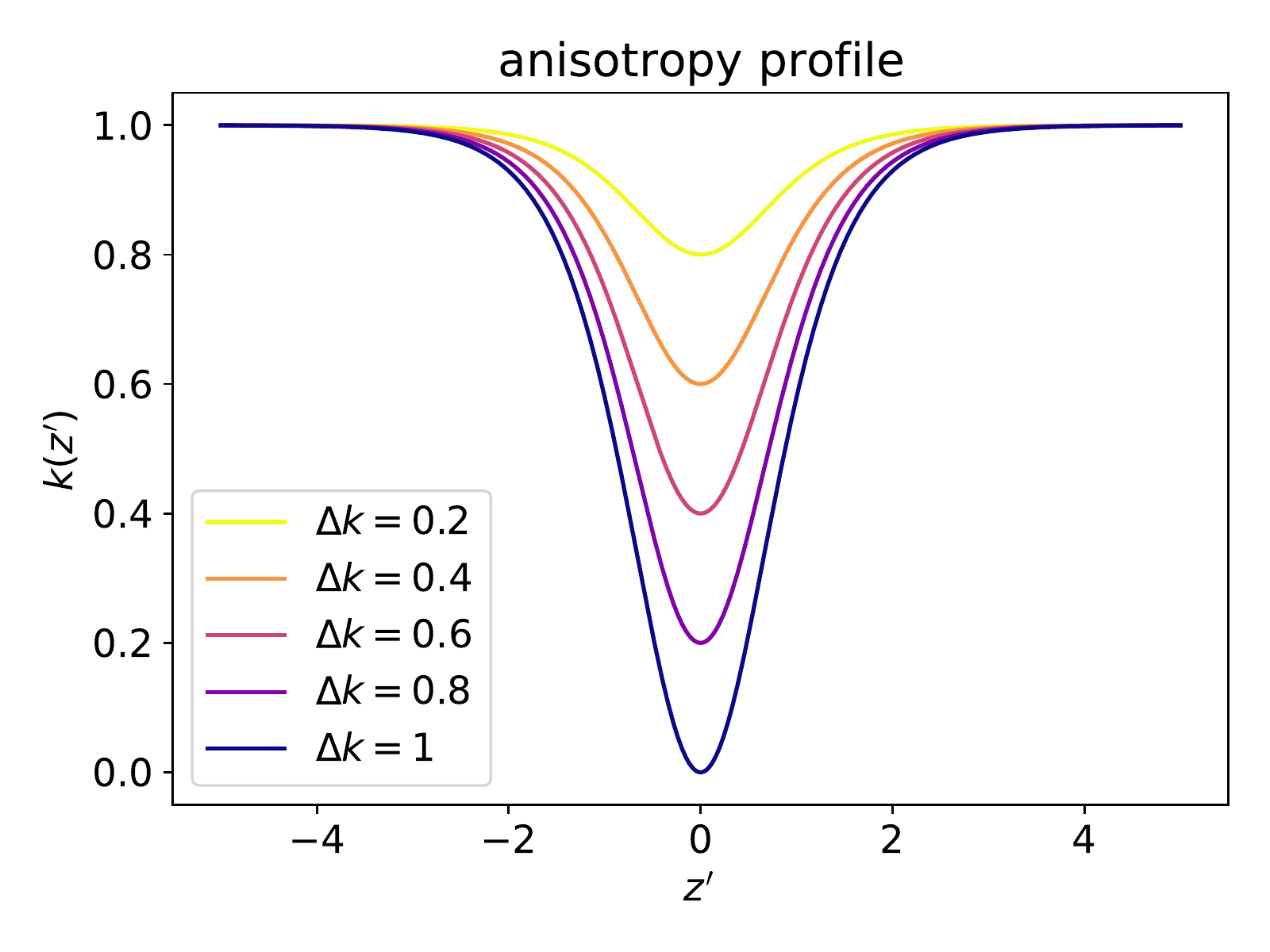}
	\caption{\label{fig:kprofile}Profile of the reduced anisotropy constant $k(z')$ in the planar defect for different values of $\Delta k$.}
\end{figure}

Variation of (\ref{eq:energy}) and assuming only small deviations of the magnetization from easy axis give the linearized micromagnetic equation 
\begin{align}
\label{eq:brown1d}
2 A\frac{\mathrm{d}^2\varphi(z)}{\mathrm{d}z^2} - \left\{2K_1(z) - \mu_0 M_\mathrm{s}H_\mathrm{ext} \right\}\varphi(z) =0. 
\end{align}
Please note that we do not explicitly account for the magnetostatic field $\mathbf{H}_\mathrm{d}$ in the linearized micromagnetic equation (\ref{eq:brown1d}). 
For many cases, the influence of the magnetostatic field $\mathbf{H}_\mathrm{d}$ may be treated by shifting the value of the external field or an effective anisotropy constant. 

We assume a defect in the magneto-crystalline anisotropy with the following profile
\begin{align}
	K_1(z) = K_1(\infty) - \frac{\Delta K}{\mathrm{cosh}^2(z/r_0)}.
\end{align} 
Introducing the Bloch parameter $\delta_0 = \sqrt{{A}/{K_1(\infty)}}$ and the following new variables
\begin{align}
	z' = \frac{z}{r_0}, \; 
	k(z) = \frac{K_1(z)}{K_1(\infty)},\; 
	h = \frac{\mu_0 H_\mathrm{ext}}{2K_1(\infty)/M_\mathrm{s}}, \;
	\bar{r}_0 = \frac{r_0}{\delta_0}
\end{align}   
we rewrite (\ref{eq:brown1d}) in dimensionless form
\begin{align}
	\label{eq:brwon1ddimless}
	\frac{\mathrm{d}^2\varphi(z')}{\mathrm{d}z'^2} - \bar{r}_0^2 \left\{k(z') - h \right\}\varphi(z') =0, \;
\end{align}
whereby the defect in reduced units is given by
\begin{align}
  k(z') = 1 - \frac{\Delta k}{\mathrm{cosh}^2(z')}, \; \Delta k = \frac{\Delta K}{K_1(\infty)}.
\end{align}
Figure \ref{fig:kprofile} shows the profile of the anisotropy defect.
Please note that the scaling of  variables is slightly different from the one used by Kronm\"uller \cite{kronmuller1987theory}. 

The nucleation field in units of $2K_1/(\mu_0 M_\mathrm{s})$ is smallest value of $h$ which solves the equation 
\begin{align}
	\label{eq:eigenproblem1d}
	\frac{\mathrm{d}^2\varphi(z')}{\mathrm{d}z'^2} + \bar{r}_0^2 \left(h-1 + \frac{\Delta k}{\mathrm{cosh}^2(z')} \right)\varphi(z') =0. 
\end{align}
Equation (\ref{eq:eigenproblem1d}) has an analytical solution similar to that of the stationary Schr\"odinger equation for the modified P\"oschl-Teller potential in one dimension \cite{dong2007factorization}. Landau and Lifshitz applied a change of variables $u = \mathrm{tanh}(z')$ for finding analytical solutions for the wave function of a particle in the potential $V = -V_0/\mathrm{cosh}^2(z)$ \cite{landau1977quantum}. This change of variables maps the problem domain from the interval $[-\infty,\infty]$ to the interval $[-1,1]$. Finally, the equation can be brought into hypergeometric form. From the condition that the solution is finite at $\pm 1$ the eigenfunction and eigenvalues were identified \cite{landau1977quantum}. The nucleation field is given by \cite{kronmuller1987theory}
\begin{align} 
  \label{eq:hnana}
  h = 1 - \frac{1}{4 \bar{r}_0^2} \left( -1 + \sqrt{1+4 \bar{r}_0^2 \Delta k } \right)^2,
\end{align}  
and the corresponding eigenfunction is \cite{landau1977quantum}
\begin{align}
 \label{eq:varphi}
 \varphi(z') = C \left( 1-\mathrm{tanh}^2(z') \right)^{\bar{r}_0 \sqrt{1-h}/2} = C \left( \frac{1}{\mathrm{cosh}(z')}\right) ^{\bar{r}_0 \sqrt{1-h}} 
\end{align}
The normalization constant $C$ is determined such that 
\begin{align}
	\int_{-\infty}^{\infty} \varphi^2(z') \mathrm{d}z' = 1.
\end{align}

\subsection{Nucleation in soft magnetic spherical inclusions}

\begin{figure}[!htb]
	\centering
	\includegraphics[scale=0.45]{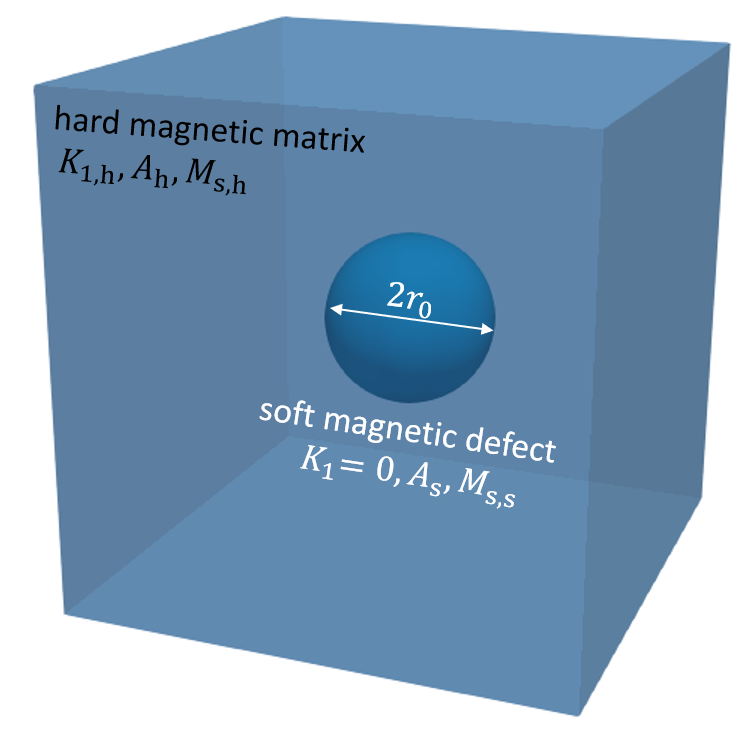}
	\caption{\label{fig:3d}Soft magnetic defect embedded within a hard magnetic materix. }
\end{figure}

Skomski and Coey \cite{skomski1993giant} studied the nucleation problem for soft magnetic spherical inclusion embedded in a hard magnetic matrix. Figure~\ref{fig:3d} shows the problem schematically. Within the sphere of diameter $2 r_0$ the magneto-crystalline anisotropy constant is zero ($K_{1,s}=0$), and the exchange constant and spontaneous magnetization might be different from the respective values in the surrounding matrix ($A_s \neq A_h, \; M_{\mathrm{s},s} \neq M_{\mathrm{s},h})$.  The subscripts $s$ and $h$ refer to the soft magnetic and hard magnetic phase, respectively. Since the solution of the corresponding eigenvalue problem is known from analogy to quantum mechanics \cite{skomski1993giant}, it is an excellent example to test conditional physics informed neural networks for a simple three-dimensional problem. 

Again we study how the magnetization deviates from the preferred crystallographic direction which coincides with the $z$ axis. All coefficients may depend on position $\mathbf{x} = (x,y,z)^\mathrm T$. For small deviations the linearized micromagnetic equation reads \cite{skomski1993giant}
\begin{align}
	\label{eq:brown3d}
	2\nabla \cdot \left( A(\mathbf{x}) \nabla \varphi(\mathbf{x}) \right) - \left\{2 K_1(\mathbf x) - \mu_0 M_\mathrm{s}(\mathbf x) H_\mathrm{ext} \right\} \varphi(\mathbf{x}) = 0.
\end{align}
If magnetostatic interactions are neglected $\varphi$ may refer to $m_x$ or $m_y$. When there is a jump of the exchange constant at the interface between the soft defect and the hard magnetic matrix the following interface condition follows from the variation of the Gibbs free energy \cite{goto1965magnetization}
\begin{align}
	\label{eq:interface}
	A_{\mathrm s} \left( \nabla \varphi_{\mathrm s} \cdot \mathbf{n} \right)= A_{\mathrm h} \left( \nabla \varphi_{\mathrm h} \cdot \mathbf{n} \right),
\end{align}
where $\mathbf{n}$ is the unit vector normal to the interface. Note that in this case we need to understand the derivatives in \eqref{eq:brown3d} in the sense of distributions \cite{henrot2006extremum} and 
the exchange constant needs to fulfill an ellipticity assumption, which is satisfied if it is uniformly bounded from below by a positive constant.  
We treat the variational formulation of the eigenvalue problem \eqref{eq:brown3d} in analogy to section~\ref{sec:varform} by a three-dimensional version of \eqref{eq:loss}.
Similar as in the one-dimensional case we introduce a new set of variables. We normalize material parameters by the respective values of the hard magnetic matrix
\begin{align}
	\tilde k(\mathbf{x}) = \frac{K_1(\mathbf{x})}{K_{1,h}}, \; \tilde a(\mathbf{x}) = \frac{A(\mathbf{x})}{A_h}, \; \tilde m(\mathbf{x}) = \frac{M_\mathbf{s}(\mathbf{x})}{M_{\mathrm s,h}}.  
\end{align}   
Outside the spherical inclusion we have $\tilde k = \tilde a = \tilde m = 1$, and inside the soft magnetic sphere we have $\tilde k = 0$.
With the Bloch parameter $\delta_{0,h} = \sqrt{{A_h}/{K_{1,h}}}$ of the hard magnetic phase and the following new variables
\begin{align}
	\mathbf{x}' = \frac{\mathbf{x}}{r_0}, \;  
	h = \frac{\mu_0 H_\mathrm{ext}}{2K_{1,h}/M_{\mathrm{s},h}}, \;
	\bar{r}_0 = \frac{r_0}{\delta_{0,h}}\\
\end{align}   
we rewrite (\ref{eq:brown3d}) in dimensionless form
\begin{align}
	\label{eq:brown3ddimless}
	\nabla' \cdot \left( \tilde a(\mathbf{x'}) \nabla' \varphi(\mathbf{x'}) \right) + \bar r_0^2 \left\{\tilde m(\mathbf{x'}) h - \tilde k(\mathbf{x'}) \right\} \varphi(\mathbf{x'}) = 0.
\end{align}
The smallest eigenvalue $h$ gives the nucleation field in units of $2K_{1,h}/M_{\mathrm{s},h}$. From the analogy of the problem with the stationary Schr\"odinger equation for a particle in the three-dimensional square well potential \cite{schiff1955quantum} the eigenvalues can be found. Taking into account the interface condition (\ref{eq:interface}) and the boundary 
condition, $\varphi \rightarrow 0$ for $|\mathbf x| \rightarrow \infty$,  Skomski and Coey \cite{skomski1993giant} derived a nonlinear equation for the lowest eigenvalue
\begin{align}
\label{eq:hn3d}
\bar r_0 \sqrt{\bar a \bar m h}\,\mathrm{cot}\left(\bar r_0 \sqrt{\frac{\bar a}{\bar m} h}\right) - \bar a + 1 + \bar r_0 \sqrt{1-h} = 0
\end{align}
with ratio of the exchange constant and the spontaneous magnetization in the defect and the matrix phase: $\bar a = A_s/A_h$ and $\bar m = M_{\mathrm s,s}/M_{\mathrm s,h}$.
 
\section{Results}

\subsection{Text book example}

We now demonstrate how to solve a simple eigenvalue problem by training  and evaluating  a dense neural network. We solve the Sturm-Liouville eigenvalue problem \eqref{eq:sturmliouville} for $p(x) = 1$, $q(x) = 0$, and $r(x) = 1$ and the boundary conditions $y(0) = 0$ and $y(1) = 0$. The lowest eigenvalue of the Sturm-Liouville eigenvalue problem for $p(x) = 1$, $q(x) = 0$, and $r(x) = 1$ and the boundary conditions $y(0) = 0$ and $y(1) = 0$ is $\lambda = \pi^2$ and the analytical solution is $y(x) = \pm sin(\pi x)$ \cite{komzsik2019applied}. 

\begin{figure}[!htb]
	\centering
	\begin{tabular}{@{}c@{}}
		\centering
		\includegraphics[scale=0.4]{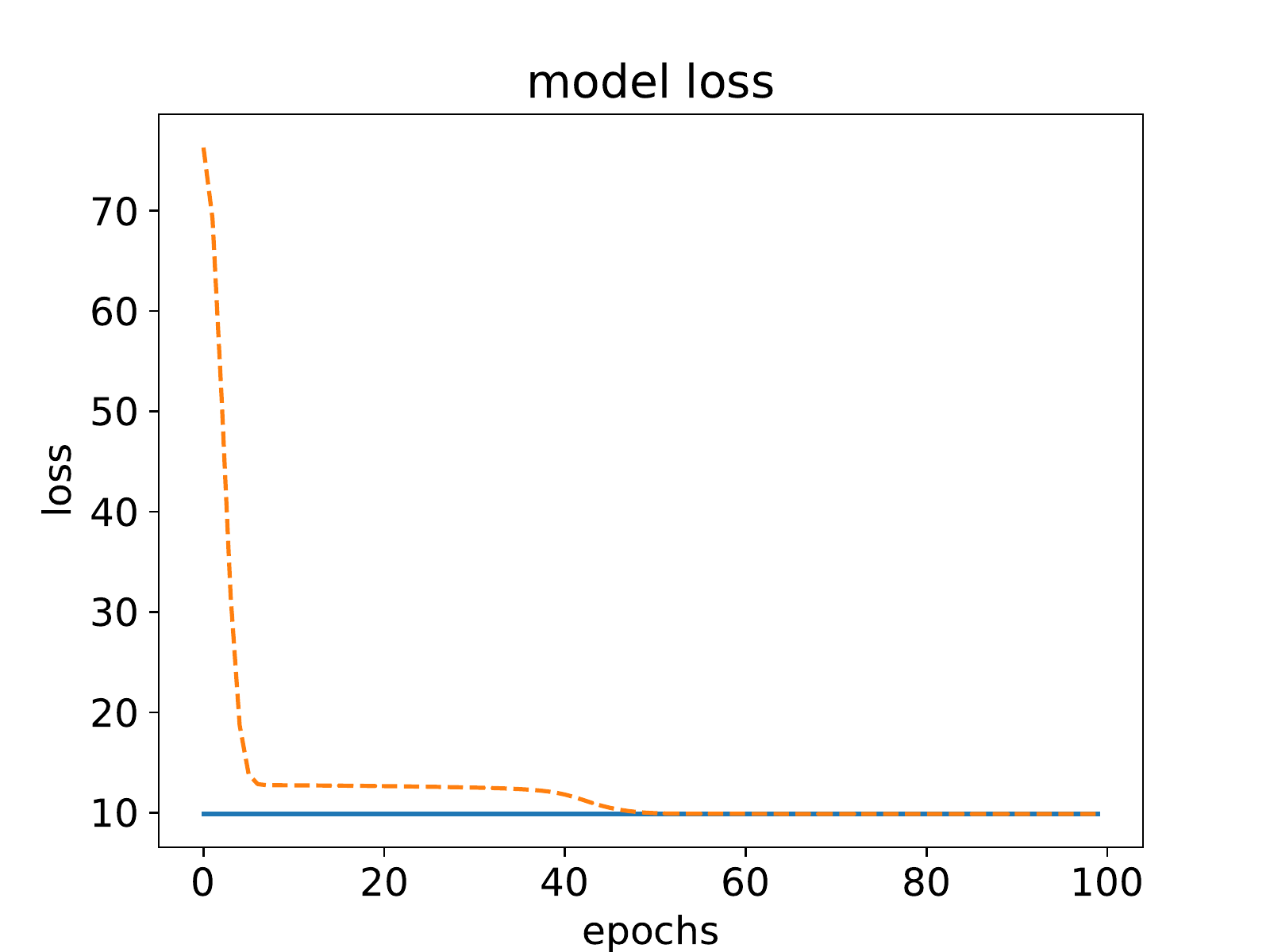}
	\end{tabular}
	\begin{tabular}{@{}c@{}}
		\centering
		\includegraphics[scale=0.4]{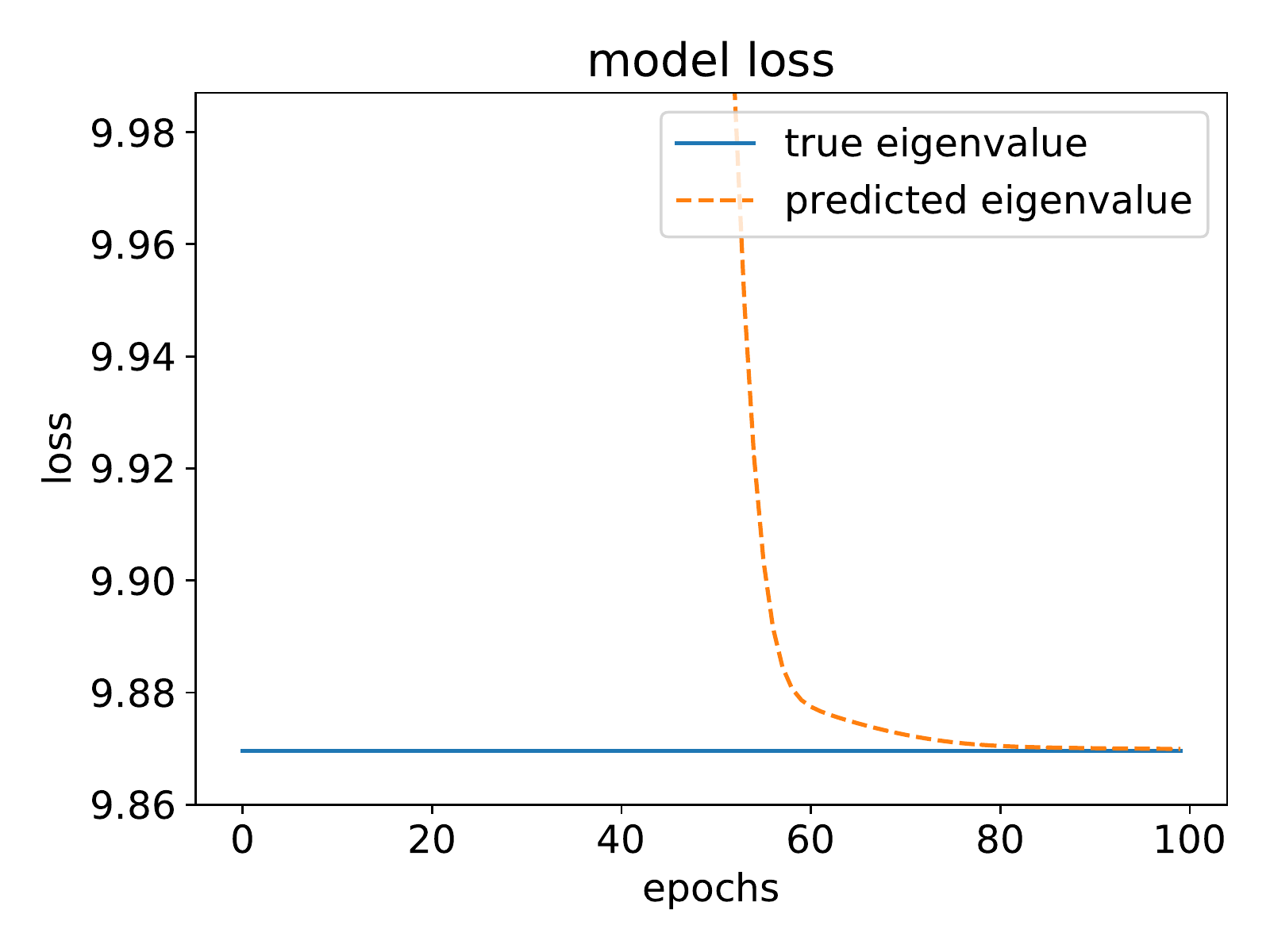}
	\end{tabular}
	\caption{Approximated eigenvalue as function of the number of complete passes through the training set (epochs). The right figure shows a zoom.}
	\label{fig:loss}
\end{figure}

\begin{figure}[!htb]
	\centering
	\includegraphics[scale=0.5]{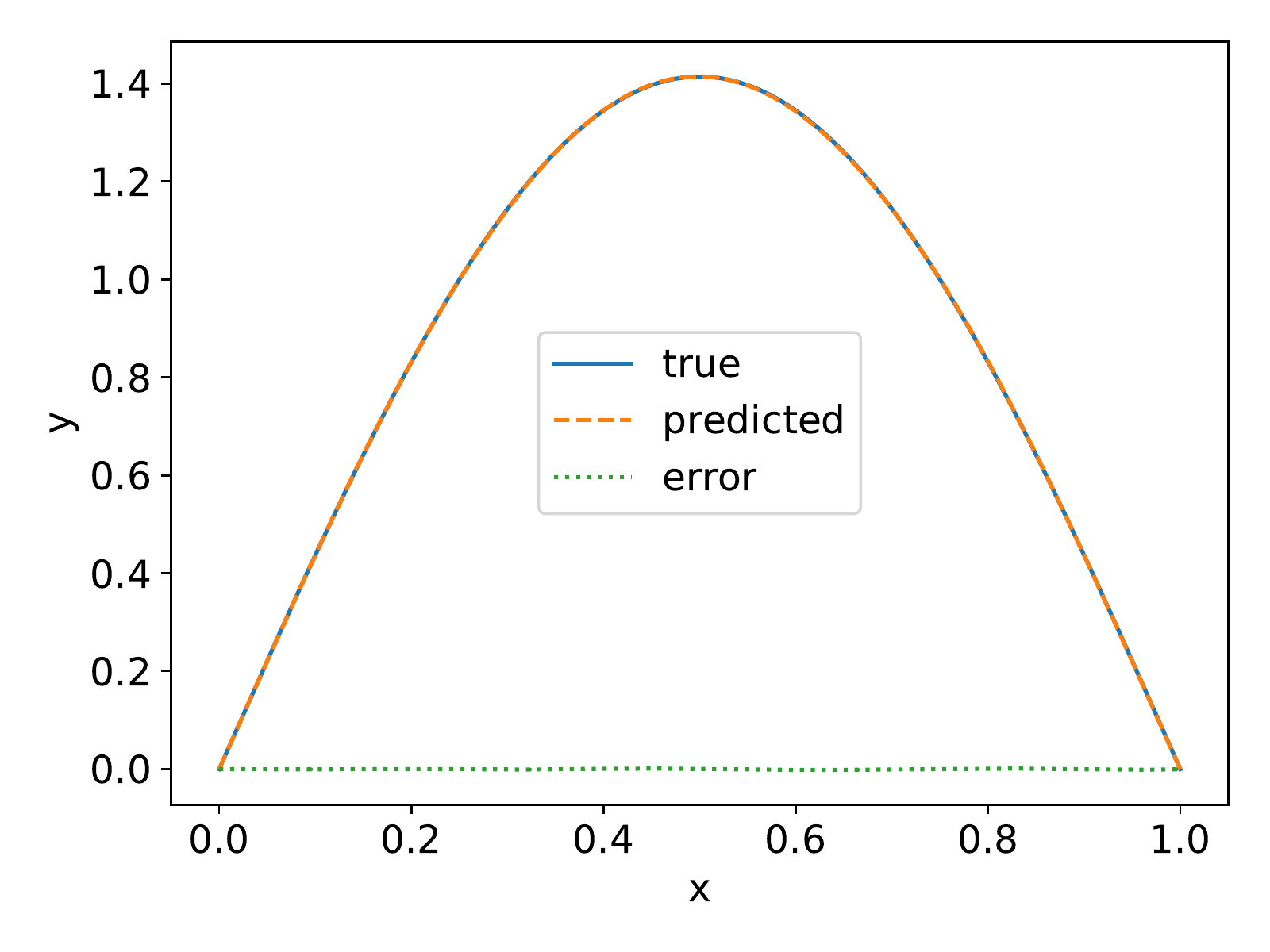}
	\caption{True and approximated eigenfunction of the test problem (Sturm-Liouville eigenvalue problem with $p(x) = 1$, $q(x) = 0$, $r(x) = 1$ and $y(0) = 0$, $y(1) = 0$).}
	\label{fig:sin}
\end{figure}

We approximate the eigenfunction with the ansatz (\ref{eq:nn_dirichlet}) using $w(x) = 1-(2x-1)^4$. For the numerical test we use a shallow network consisting of two hidden layers with four neurons each. We used the hyperbolic tangent as activation function. Training was done using the Adam algorithm \cite{kingma2014adam} with an initial step size of $10^{-4}$. The batch size was $N=2^{10}$ and we used a total of $2^8$ batches for training.  The number of complete passes through the training set (epochs) was 100. We used $\gamma = 64$ for the prefactor of the penalty term.

Figure \ref{fig:loss} shows how the function (\ref{eq:loss}) decreases with the number of epochs. Please note that after training the value $L(y_\mathrm{approx}(x))$ approximates the lowest eigenvalue because the penalty term vanishes if the training was successful. The absolute error between the approximated and true eigenvalue is $0.000175$. 

Figure \ref{fig:sin} compares the true and approximated eigenfunction.  

\subsection{Inhomogeneous nucleation in magnetic defects}

We now want to estimate the nucleation field for a class of magnets with a defect. We train a neural network for the deviation $\varphi$ of the magnetization from its equilibrium position for defects characterized by the defect width and the defect strength. At the surface of the magnet the normal derivative of $\varphi$ needs to be zero. This natural boundary condition arises from the variation of the Gibbs free energy of the system \cite{brown1963micromagnetics}.

\begin{figure}[!htb]
	\centering
	\includegraphics[scale=0.5]{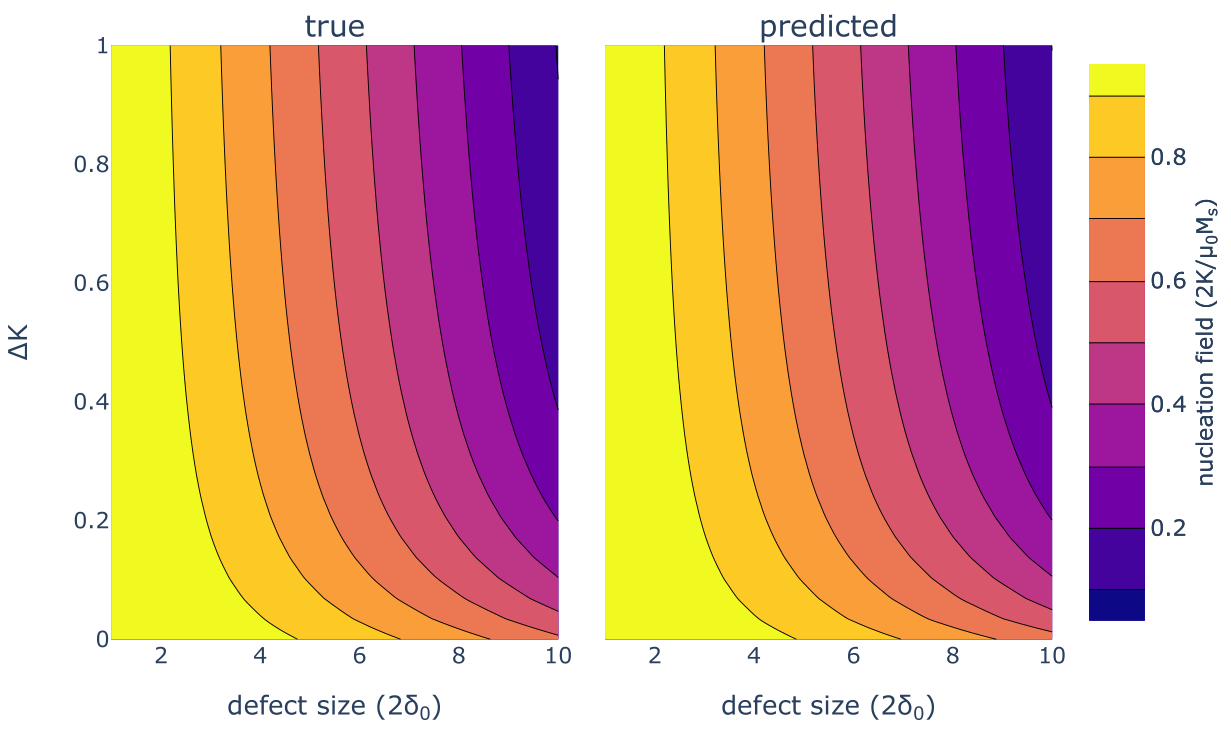}
	\caption{\label{fig:hn}True and predicted values for the nucleation field as a function of the defect size and defect strength for a network layout with three hidden layers with eight neurons each. The defect size is given in units of $2 \delta_0$, where $\delta_0$ is the Bloch parameter of the hard magnetic matrix.}
\end{figure}

\begin{figure}[!htb]
	\centering
	\includegraphics[scale=0.5]{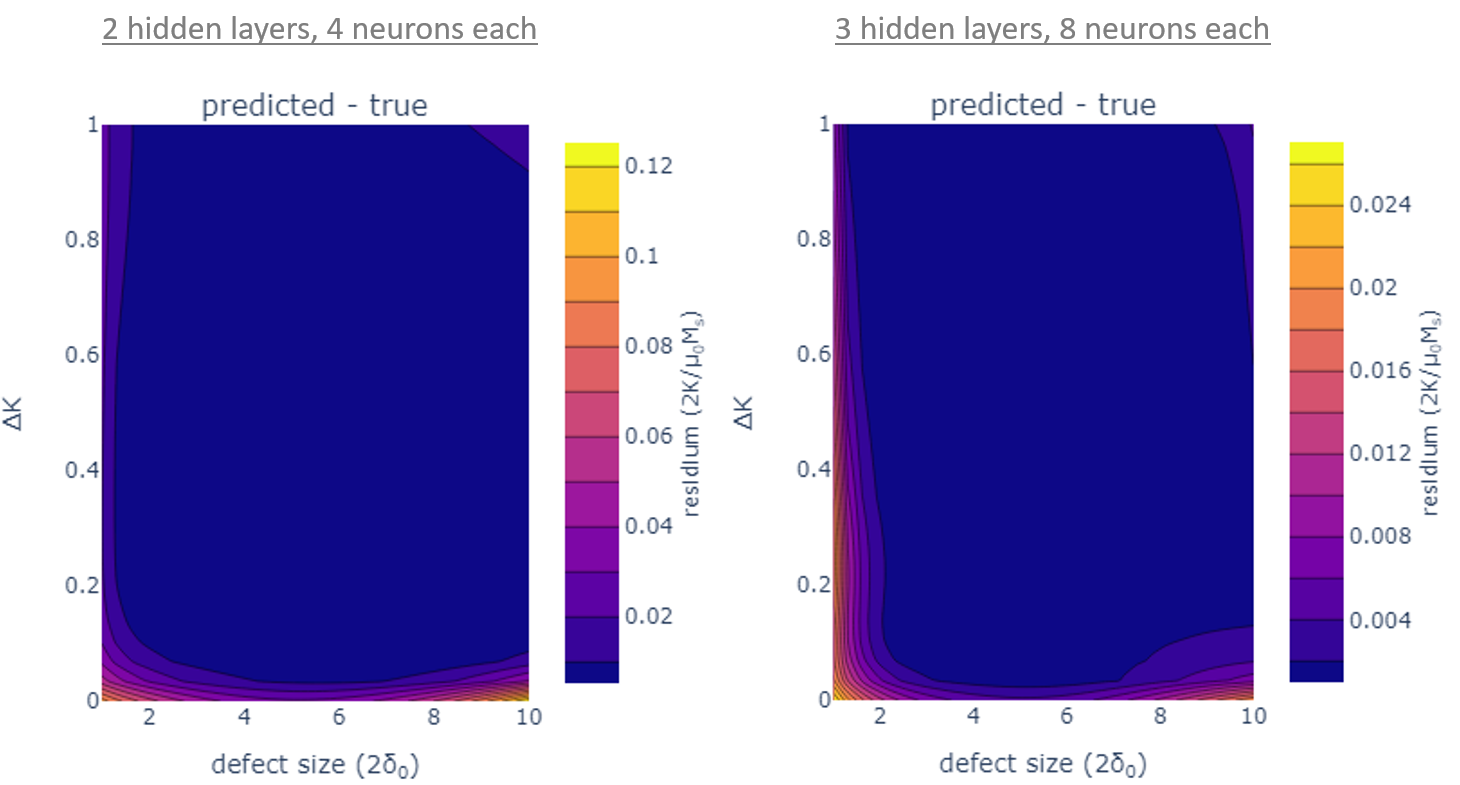}
	\caption{\label{fig:reshn}Residuum for the nucleation field as function of defect size and defect strength computed for the 2x[4] network and the 3x[8] network. The defect size is given in units of $2 \delta_0$, where $\delta_0$ is the Bloch parameter of the hard magnetic matrix.}
\end{figure}

\begin{figure}[!htb]
	\centering
	\includegraphics[scale=0.40]{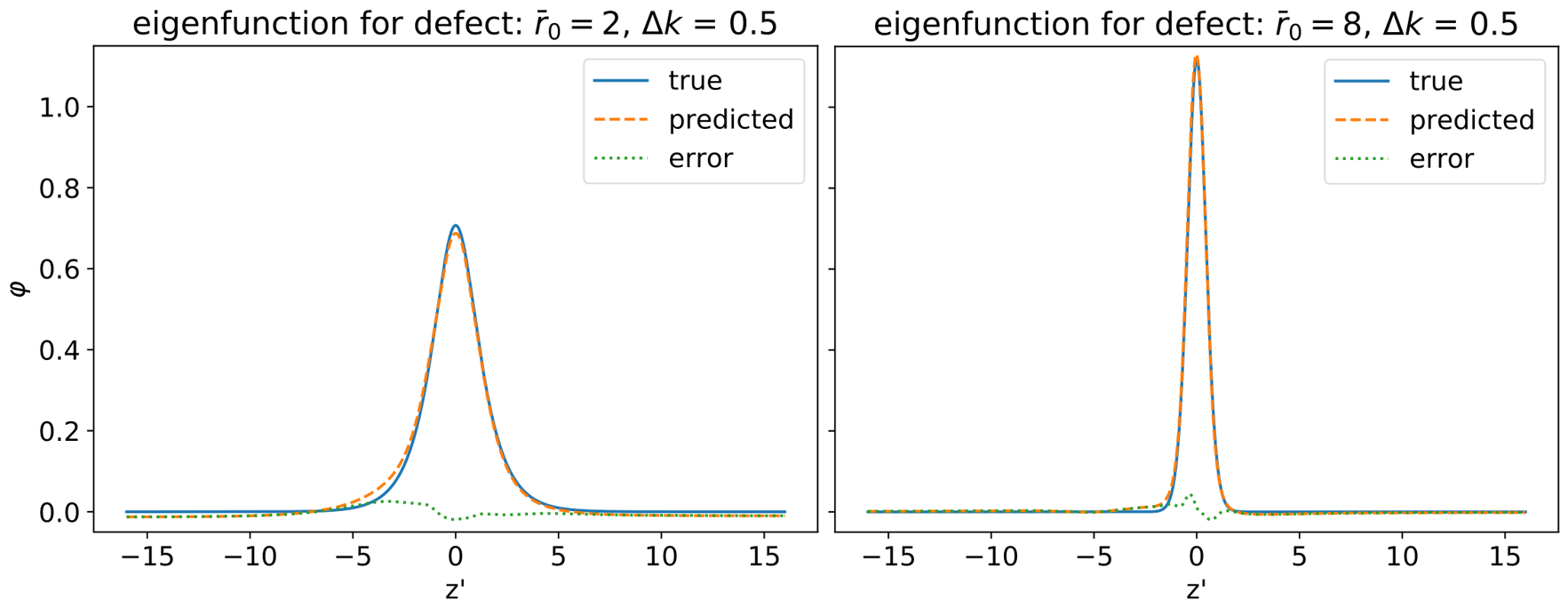}
	\caption{\label{fig:phi}True and approximated eigenfunctions for inhomogeneous nucleation  in planar defects. The functions show how the magnetization deviates from the equilibrium position for defects of width $2\bar{r}_0$ and strength $\Delta k$ when the external field reaches the nucleation field. The network layout was 3x[8].}
\end{figure}

In order to compare the numerical result with the analytical solution of (\ref{eq:eigenproblem1d}), we simply truncate the problem domain and apply natural boundary conditions at $z' = \pm 16$. Please note that the defect size in reduced units  has a width of 2 and that we expect $\varphi$ to be of similar shape as the profile of the magneto-crystalline anisotropy constant plotted in Figure~\ref{fig:kprofile}. 

For solving (\ref{eq:eigenproblem1d}) we set the coefficients of the Sturm-Liouville eigenvalue problem to $p = 1/\bar r_0^2$, $q(z') = -1+\Delta k/\cosh^2(z')$, and $r=1$. The eigenfunction, which describes how the magnetization will start to move out from the anisotropy direction, when the external field reaches the nucleation field is approximated by a neural network
\begin{align}
	\label{eq:phinn}
	\varphi_\mathrm{approx}(z') = \mathcal N(z',\mathbf{w};\bar{r}_0,\Delta k)
\end{align}
The input for the network is the position $z'$ within the magnet and the tags which define the defect width, $2 \bar{r}_0$, and the strength of the defect, $\Delta k$. The tags are the conditional input (see Figure \ref{fig:cpinn}) which selects a specific defect. Thus, we can learn a set of solutions for the class of problems described by the parameters $\bar{r}_0$ and $\Delta k$. 
 
During training of the network we adjust the weights $\mathbf{w}$ such that the functional \eqref{eq:loss} is minimized. For training, we use a batch size $N = 2^{12}$. For all samples within a batch, $z'$ varies but the parameters $\bar{r}_0$ and $\Delta k$ are kept constant.
$M = 2^{10}$ pairs $(\bar{r}_0, \Delta k)$ are quasi-randomly sampled using a Sobol sequence from the set $\{(\bar{r}_0,\Delta k)\,|\,1 \le \bar{r}_0 \le 10, 0 \le \Delta k \le 1\}$. Batches with the same tag vector are repeated $I=4$ times. For training, we applied the Adam method \cite{kingma2014adam} with an initial step length of $10^{-3}$.
Too large values for the prefactor $\gamma$ of the penalty term slows down convergence. Thus, we set $\gamma = 1$. 
 
We compared the performance for two different layouts of the neural network. One with two hidden layers made of four neurons each (2x[4])  and another with three hidden layers made of eight neurons each (3x[8]). The mean absolute error in the nucleation field computed over 900 regularly distributed points in 
the $(\bar{r}_0,\Delta k)$ - plane was $0.006$ and $0.002$ for the 2x[4] and 3x[8] network, respectively.  Figure~\ref{fig:hn} shows the contour plot of the nucleation field in the $(\bar r_0,\Delta k)$ - plane. The left hand side shows the analytic solution for $h_\mathrm{true}$ according to (\ref{eq:hnana}), the right hand side gives the eigenvalue $h_\mathrm{predicted}$ computed with the 3x[8] network. 
The residua $h_\mathrm{predicted}-h_\mathrm{true}$ for the two different network layouts are given in  Figure~\ref{fig:reshn}. 

We also compare the true and predicted eigenfunctions for selected values of the tags  
$(\bar r_0, \;\Delta k)$. Figure~\ref{fig:phi} compares $\varphi(z')$ as given by equations (\ref{eq:varphi}) and (\ref{eq:phinn}), respectively.

\subsection{Estimating nucleation fields for three-dimensional problems} 
\label{sec:3d}
Finally, we apply the proposed approach to solve a micromagnetic problem in three dimensions. We train a neural network to predict the nucleation field of a spherical soft magnetic phase embedded in a hard magnetic matrix. Our goal is to predict the nucleation field depending on size of the inclusion, the magnetization of the soft magnetic phase, and the exchange constant of the soft magnetic phase. We will use $(\bar r_0, \tilde m, \; \tilde a)$ as tags of a conditional physics informed neural network that estimates the solution of the eigenproblem (\ref{eq:brown3ddimless}). 
We minimize the three-dimensional extension of (\ref{eq:loss}) with $p(\mathbf{x'}) = \tilde a(\mathbf{x'})/\bar r_0^2$, $q(\mathbf{x'}) = -\tilde k(\mathbf{x'})$, and $r(\mathbf{x'}) = \tilde m (\mathbf{x'})$. 
Note that the non-smoothness of the exchange constant is included in our setting of section~\ref{sec:varform}.
As in the one-dimensional case, we truncate the problem domain and apply natural boundary conditions for $\varphi$.   
The transverse component that shows how the magnetization starts to deviate from the anisotropy direction, when the external field reaches the nucleation field is approximated with
\begin{align}
		\varphi_\mathrm{approx}(\mathbf x') = \mathcal N(\mathbf x',\mathbf{w};\bar{r}_0,\tilde m, \tilde a).
\end{align}  

\begin{figure}[!htb]
	\centering
	\includegraphics[scale=0.69]{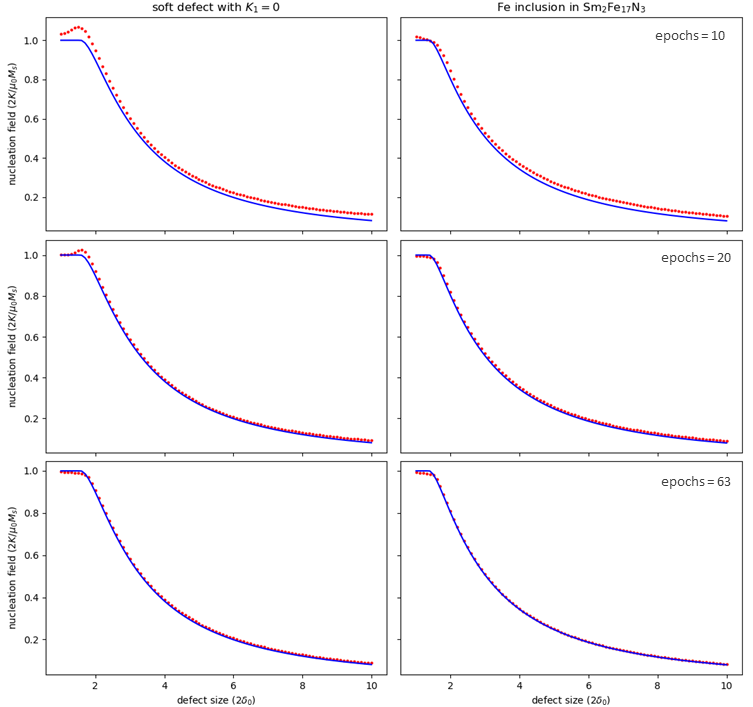}
	\caption{\label{fig:hn3d}True (solid lines) and predicted values (symbols) for the nucleation field as function of defect size for two different soft magnetic phases. Left: $(\tilde m = 1,\; \tilde a = 1)$; Right: $(\tilde m = 1.387\; \tilde a = 1.5)$. The defect size is given in units of $2 \delta_0$, where $\delta_0$ is the Bloch parameter of the hard magnetic matrix. From top to bottom the number of epochs are 10, 20, and 63.}
\end{figure}

We sample the training points from the box $(-5,5)^3$ using a Sobol sequence. Typically, soft magnetic phases show a higher magnetization and higher exchange constant than hard magnetic phases. Therefore, we chose $\tilde m \ge 1$ and $\tilde a \ge 1$. In order to train the network for a variety of soft inclusions, we sample $M = 2^{10}$ triplets from 
the set $\{(\bar r_0, \tilde m, \tilde a) \,|\, 1 \le \bar r_0 \le 10,\; 1 \le \tilde m \le \sqrt{2},\; 1 \le \tilde a \le 2 \}$ using a Sobol sequence. By setting the upper bounds we take into account that the exchange constant may scale with the saturation magnetization squared \cite{kronmuller2003micromagnetism}. For each batch the tags are kept the same. We set the batch size $N = 2^{12}$. Batches with the same tag vector are repeated $I=2^4$ times. The penalty parameter is $\gamma = 1$. The neural network consists of 5 hidden layers with 8 neurons each. For training, we applied the Adam method \cite{kingma2014adam} with an initial step length (learning rate) of $10^{-3}$. 

For comparison of the nucleation field estimate with its theoretical value we solved (\ref{eq:hn3d}) numerically, using the Powell hybrid method as implemented in the Python library SciPy \cite{2020SciPy-NMeth}. Figure~\ref{fig:hn3d} shows the nucleation field as function of the defect size $\bar r_0$ for two different magnetic phases in the defect. The curve for $\tilde m = 1.387$ and $\tilde a = 1.5$ corresponds to an Fe inclusion in a Sm$_2$Fe$_{17}$N$_3$ matrix \cite{skomski1993giant}. Please note, that the normalized nucleation field $h$ is constant for small radii of the spherical inclusion: $h = 1$ for $\bar r_0 \lesssim 1$ \cite{skomski1999}. In Figure~\ref{fig:hn3d} shows how the neural network estimate approaches the true solution with increasing number of epochs. The lowest validation loss was found for $\mathrm{epoch}=63$.  The largest errors occur next to the knee in the $h(\bar r_0)$ curve. The mean absolute errors are 0.007 and 0.004 for the  defect with $K_1 = 0$ and for the Fe inclusion in  Sm$_2$Fe$_{17}$N$_3$, respectively. The mean absolute erros were computed for the 91 points shown in Figure~\ref{fig:hn3d}. 

\section{Conclusion}

We proposed to use conditional physics informed neural networks to solve classes of eigenvalue problems. The methodology follows previous work on using neural networks for solving variational problems \cite{e2018deep}. The unknown eigenfunction is approximated by a neural network. Its weights are found by minimizing a loss function that is closely related to the Rayleigh Ritz coefficient. Inputs to the network are points sampled in the problem domain. In our modification of the method, parameters that determine the coefficients of the eigenvalue equation serve as additional input. During training the network learns how eigenfunction and eigenvalue depend on the parameters. When the network is used for prediction, the parameters serve as condition that selects a specific solution out of a set of eigenvalue problems.

We demonstrated the proposed method for classical eigenvalue problems in micromagnetics. We speculate that the method can be used in the future to quickly approximate solutions of eigenvalue problems in engineering. Especially, we envision to use conditional physics informed neural networks for permanent magnet design. Permanent magnets are essential technologies for low-carbon power generation and low-carbon transport. For magnets made of core-shell grains \cite{ito2016coercivity}, geometry optimization \cite{skomski2014geometry} and materials selection rely heavily on the fast approximation of coercivity. An unsupervised training neural network for coercive field evaluation may serve as a basic building block for an active learning scheme \cite{balachandran2019machine} for magnetic materials development.

\section*{Acknowledgment}
The financial support by the Austrian Federal Ministry for Digital and Economic  Affairs, the National Foundation for Research, Technology and Development and the Christian Doppler Research Association is gratefully acknowledged. 
L.E. acknowledges support by the Austrian Science Foundation (FWF) under grant No. P31140-N32.

\bibliographystyle{elsarticle-num-names}
\bibliography{references.bib}
\end{document}